# Deep Learning vs. Traditional Computer Vision

Niall O' Mahony, Sean Campbell, Anderson Carvalho, Suman Harapanahalli,
Gustavo Velasco Hernandez, Lenka Krpalkova, Daniel Riordan, Joseph Walsh

IMaR Technology Gateway, Institute of Technology Tralee, Tralee, Ireland
`niall.omahony@research.ittralee.ie`

**Abstract.** Deep Learning has pushed the limits of what was possible in the domain of Digital Image Processing. However, that is not to say that the traditional computer vision techniques which had been undergoing progressive development in years prior to the rise of DL have become obsolete. This paper will analyse the benefits and drawbacks of each approach. The aim of this paper is to promote a discussion on whether knowledge of classical computer vision techniques should be maintained. The paper will also explore how the two sides of computer vision can be combined. Several recent hybrid methodologies are reviewed which have demonstrated the ability to improve computer vision performance and to tackle problems not suited to Deep Learning. For example, combining traditional computer vision techniques with Deep Learning has been popular in emerging domains such as Panoramic Vision and 3D vision for which Deep Learning models have not yet been fully optimised.

## 1 Introduction

Deep Learning (DL) is used in the domain of digital image processing to solve difficult problems (e.g. image colourization, classification, segmentation and detection). DL methods such as Convolutional Neural Networks (CNNs) mostly improve prediction performance using big data and plentiful computing resources and have pushed the boundaries of what was possible. Problems which were assumed to be unsolvable are now being solved with super-human accuracy. Image classification is a prime example of this. Since being reignited by Krizhevsky, Sutskever and Hinton in 2012 [1], DL has dominated the domain ever since due to a substantially better performance compared to traditional methods .

Is DL making traditional Computer Vision (CV) techniques obsolete? Has DL superseded traditional computer vision? Is there still a need to study traditional CV techniques when DL seems to be so effective? These are all questions which have been brought up in the community in recent years [2], which this paper intends to address.

Additionally, DL is not going to solve all CV problems. There are some problems where traditional techniques with global features are a better solution. The advent of DL may open many doors to do something with traditional techniques to overcome the many challenges DL brings (e.g. computing power, time, accuracy, characteristics and quantity of inputs, and among others).

This paper will provide a comparison of deep learning to the more traditional hand-crafted feature definition approaches which dominated CV prior to it. There has been so much progress in Deep Learning in recent years that it is impossible for this paper to capture the many facets and sub-domains of Deep Learning which are tackling the most pertinent problems in CV today. This paper will review traditional algorithmic approaches in CV, and more particularly, the applications in which they have been used as an adequate substitute for DL, to complement DL and to tackle problems DL cannot.

The paper will then move on to review some of the recent activities in combining DL with CV, with a focus on the state-of-the-art techniques for emerging technology such as 3D perception, namely object registration, object detection and semantic segmentation of 3D point clouds. Finally, developments and possible directions of getting the performance of 3D DL to the same heights as 2D DL are discussed along with an outlook on the impact the increased use of 3D will have on CV in general.

## 2 A Comparison of Deep Learning and Traditional Computer Vision

### 2.1 What is Deep Learning

To gain a fundamental understanding of DL we need to consider the difference between descriptive analysis and predictive analysis.

Descriptive analysis involves defining a comprehensible mathematical model which describes the phenomenon that we wish to observe. This entails collecting data about a process, forming hypotheses on patterns in the data and validating these hypotheses through comparing the outcome of descriptive models we form with the real outcome [3]. Producing such models is precarious however because there is always a risk of un-modelled variables that scientists and engineers neglect to include due to ignorance or failure to understand some complex, hidden or non-intuitive phenomena [4].

Predictive analysis involves the discovery of rules that underlie a phenomenon and form a predictive model which minimise the error between the actual and the predicted outcome considering all possible interfering factors [3]. Machine learning rejects the traditional programming paradigm where problem analysis is replaced by a training framework where the system is fed a large number of training patterns (sets of inputs for which the desired outputs are known) which it learns and uses to compute new patterns [5].

DL is a subset of machine learning. DL is based largely on Artificial Neural Networks (ANNs), a computing paradigm inspired by the functioning of the human brain. Like the human brain, it is composed of many computing cells or 'neurons' that each perform a simple operation and interact with each other to make a decision [6]. Deep Learning is all about learning or 'credit assignment' across many layers of a neural network accurately, efficiently and without supervision and is of recent interest due to enabling advancements in processing hardware [7]. Self-organisation and the exploitation of interactions between small units have proven to perform better than central control, particularly for complex non-linear process models in that better fault tolerance and adaptability to new data is achievable [7].

## 2.2 Advantages of Deep Learning

Rapid progressions in DL and improvements in device capabilities including computing power, memory capacity, power consumption, image sensor resolution, and optics have improved the performance and cost-effectiveness of further quickened the spread of vision-based applications. Compared to traditional CV techniques, DL enables CV engineers to achieve greater accuracy in tasks such as image classification, semantic segmentation, object detection and Simultaneous Localization and Mapping (SLAM). Since neural networks used in DL are trained rather than programmed, applications using this approach often require less expert analysis and fine-tuning and exploit the tremendous amount of video data available in today's systems. DL also provides superior flexibility because CNN models and frameworks can be re-trained using a custom dataset for any use case, contrary to CV algorithms, which tend to be more domain-specific.

Taking the problem of object detection on a mobile robot as an example, we can compare the two types of algorithms for computer vision:

The traditional approach is to use well-established CV techniques such as feature descriptors (SIFT, SURF, BRIEF, etc.) for object detection. Before the emergence of DL, a step called feature extraction was carried out for tasks such as image classification. Features are small "interesting", descriptive or informative patches in images. Several CV algorithms, such as edge detection, corner detection or threshold segmentation may be involved in this step. As many features as practicable are extracted from images and these features form a definition (known as a bag-of-words) of each object class. At the deployment stage, these definitions are searched for in other images. If a significant number of features from one bag-of-words are in another image, the image is classified as containing that specific object (i.e. chair, horse, etc.).

The difficulty with this traditional approach is that it is necessary to choose which features are important in each given image. As the number of classes to classify increases, feature extraction becomes more and more cumbersome. It is up to the CV engineer's judgment and a long trial and error process to decide which features best describe different classes of objects. Moreover, each feature definition requires dealing with a plethora of parameters, all of which must be fine-tuned by the CV engineer.

DL introduced the concept of end-to-end learning where the machine is just given a dataset of images which have been annotated with what classes of object are present in each image [7]. Thereby a DL model is 'trained' on the given data, where neural networks discover the underlying patterns in classes of images and automatically works out the most descriptive and salient features with respect to each specific class of object for each object. It has been well-established that DNNs perform far better than traditional algorithms, albeit with trade-offs with respect to computing requirements and training time. With all the state-of-the-art approaches in CV employing this methodology, the workflow of the CV engineer has changed dramatically where the knowledge and expertise in extracting hand-crafted features has been replaced by knowledge and expertise in iterating through deep learning architectures as depicted in Fig. 1.

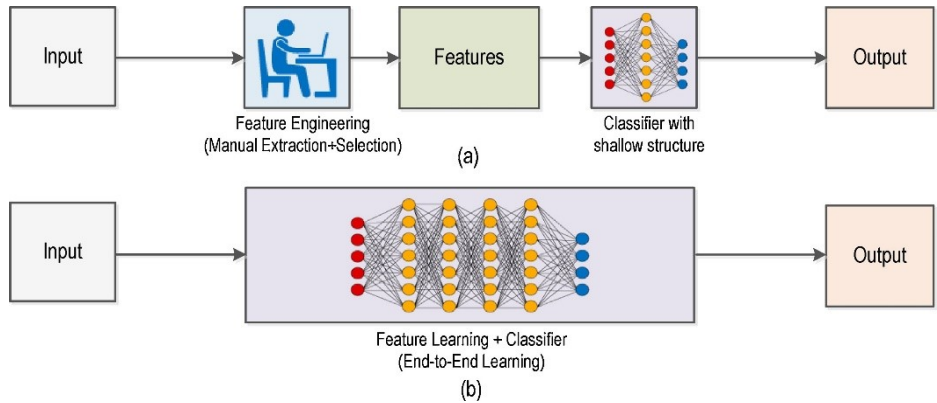

**Fig. 1.** (a) Traditional Computer Vision workflow vs. (b) Deep Learning workflow. Figure from [8].

The development of CNNs has had a tremendous influence in the field of CV in recent years and is responsible for a big jump in the ability to recognize objects [9]. This burst in progress has been enabled by an increase in computing power, as well as an increase in the amount of data available for training neural networks. The recent explosion in and wide-spread adoption of various deep-neural network architectures for CV is apparent in the fact that the seminal paper ImageNet Classification with Deep Convolutional Neural Networks has been cited over 3000 times [2].

CNNs make use of kernels (also known as filters), to detect features (e.g. edges) throughout an image. A kernel is just a matrix of values, called weights, which are trained to detect specific features. As their name indicates, the main idea behind the CNNs is to spatially convolve the kernel on a given input image check if the feature it is meant to detect is present. To provide a value representing how confident it is that a specific feature is present, a convolution operation is carried out by computing the dot product of the kernel and the input area where kernel is overlapped (the area of the original image the kernel is looking at is known as the receptive field [10]).

To facilitate the learning of kernel weights, the convolution layer's output is summed with a bias term and then fed to a non-linear activation function. Activation Functions are usually non-linear functions like Sigmoid, TanH and ReLU (Rectified Linear Unit). Depending on the nature of data and classification tasks, these activation functions are selected accordingly [11]. For example, ReLUs are known to have more biological representation (neurons in the brain either fire or they don't). As a result, it yields favourable results for image recognition tasks as it is less susceptible to the vanishing gradient problem and it produces sparser, more efficient representations [7].

To speed up the training process and reduce the amount of memory consumed by the network, the convolutional layer is often followed by a pooling layer to remove redundancy present in the input feature. For example, max pooling moves a window over the input and simply outputs the maximum value in that window effectively reducing to the important pixels in an image [7]. As shown in Fig. 2, deep CNNs may have several pairs of convolutional and pooling layers. Finally, a Fully Connected layer flattens the previous layer volume into a feature vector and then an output layer which computes the scores (confidence or probabilities) for the output classes/features through a dense network. This output is then passed to a regression function such as Softmax [12], for example, which maps everything to a vector whose elements sum up to one [7].

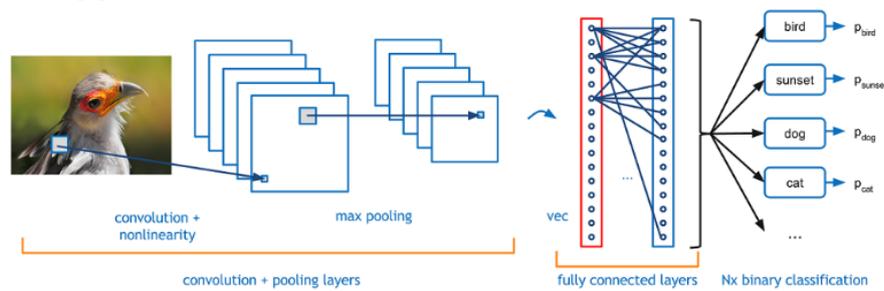

**Fig. 2.** Building blocks of a CNN. Figure from [13]

But DL is still only a tool of CV For example, the most common neural network used in CV is the CNN. But what is a convolution? It's in fact a widely used image processing technique (e.g. see Sobel edge detection). The advantages of DL are clear, and it would be beyond the scope of this paper to review the state-of-the-art. DL is certainly not the panacea for all problems either, as we will see in following sections of this paper, there are problems and applications where the more conventional CV algorithms are more suitable.

### 2.3 Advantages of Traditional Computer Vision Techniques

This section will detail how the traditional feature-based approaches such as those listed below have been shown to be useful in improving performance in CV tasks:
- Scale Invariant Feature Transform (SIFT) [14]
- Speeded Up Robust Features (SURF) [15]
- Features from Accelerated Segment Test (FAST) [16]
- Hough transforms [17]
- Geometric hashing [18]

Feature descriptors such as SIFT and SURF are generally combined with traditional machine learning classification algorithms such as Support Vector Machines and K-Nearest Neighbours to solve the aforementioned CV problems.

DL is sometimes overkill as often traditional CV techniques can solve a problem much more efficiently and in fewer lines of code than DL. Algorithms like SIFT and even simple colour thresholding and pixel counting algorithms are not class-specific, that is, they are very general and perform the same for any image. In contrast, features learned from a deep neural net are specific to your training dataset which, if not well constructed, probably won't perform well for images different from the training set. Therefore, SIFT and other algorithms are often used for applications such as image stitching/3D mesh reconstruction which don't require specific class knowledge. These tasks have been shown to be achievable by training large datasets, however this requires a huge research effort and it is not practical to go through this effort for a closed application. One needs to practice common sense when it comes to choosing which route to take for a given CV application. For example, to classify two classes of product on an assembly line conveyor belt, one with red paint and one with blue paint. A deep neural net will work given that enough data can be collected to train from. However, the same can be achieved by using simple colour thresholding. Some problems can be tackled with simpler and faster techniques.

What if a DNN performs poorly outside of the training data? If the training dataset is limited, then the machine may overfit to the training data and not be able to generalize for the task at hand. It would be too difficult to manually tweak the parameters of the model because a DNN has millions of parameters inside of it each with complex inter-relationships. In this way, DL models have been criticised to be a black box in this way [5]. Traditional CV has full transparency and the one can judge whether your solution will work outside of a training environment. The CV engineer can have insights into a problem that they can transfer to their algorithm and if anything fails, the parameters can be tweaked to perform well for a wider range of images.

Today, the traditional techniques are used when the problem can be simplified so that they can be deployed on low cost microcontrollers or to limit the problem for deep learning techniques by highlighting certain features in data, augmenting data [19] or aiding in dataset annotation [20]. We will discuss later in this paper how many image transformation techniques can be used to improve your neural net training. Finally, there are many more challenging problems in CV such as: Robotics [21], augmented reality [22], automatic panorama stitching [23], virtual reality [24], 3D modelling [24], motion estimation [24], video stabilization [21], motion capture [24], video processing [21] and scene understanding [25] which cannot simply be easily implemented in a differentiable manner with deep learning but benefit from solutions using "traditional" techniques.

## 3   Challenges for Traditional Computer Vision

### 3.1   Mixing Hand-Crafted Approaches with DL for Better Performance

There are clear trade-offs between traditional CV and deep learning-based approaches. Classic CV algorithms are well-established, transparent, and optimized for performance and power efficiency, while DL offers greater accuracy and versatility at the cost of large amounts of computing resources.

Hybrid approaches combine traditional CV and deep learning and offer the advantages traits of both methodologies. They are especially practical in high performance systems which need to be implemented quickly. For example, in a security camera, a CV algorithm can efficiently detect faces or other features [26] or moving objects [27] in the scene. These detections can then be passed to a DNN for identity verification or object classification. The DNN need only be applied on a small patch of the image saving significant computing resources and training effort compared to what would be required to process the entire frame.

The fusion of Machine Learning metrics and Deep Network have become very popular, due to the simple fact that it can generate better models. Hybrid vision processing implementations can introduce performance advantage and 'can deliver a 130X-1,000X reduction in multiply-accumulate operations and about 10X improvement in frame rates compared to a pure DL solution. Furthermore, the hybrid implementation uses about half of the memory bandwidth and requires significantly lower CPU resources' [28].

## 3.2   Overcoming the Challenges of Deep Learning

There are also challenges introduced by DL. The latest DL approaches may achieve substantially better accuracy; however this jump comes at the cost of billions of additional math operations and an increased requirement for processing power. DL requires a these computing resources for training and to a lesser extent for inference. It is essential to have dedicated hardware (e.g. high-powered GPUs[29] and TPUs [30] for training and AI accelerated platforms such as VPUs for inference [31]) for developers of AI.

Vision processing results using DL are also dependent on image resolution. Achieving adequate performance in object classification, for example, requires high-resolution images or video – with the consequent increase in the amount of data that needs to be processed, stored, and transferred. Image resolution is especially important for applications in which it is necessary to detect and classify objects in the distance, e.g. in security camera footage. The frame reduction techniques discussed previously such as using SIFT features [26, 32] or optical flow for moving objects [27] to first identify a region of interest are useful with respect to image resolution and also with respect to reducing the time and data required for training.

DL needs big data. Often millions of data records are required. For example, PASCAL VOC Dataset consists of 500K images with 20 object categories [26][33], ImageNet consists of 1.5 million images with 1000 object categories [34] and Microsoft Common Objects in Context (COCO) consists of 2.5 million images with 91 object categories [35]. When big datasets or high computing facility are unavailable, traditional methods will come into play.

Training a DNN takes a very long time. Depending on computing hardware availability, training can take a matter of hours or days. Moreover, training for any given application often requires many iterations as it entails trial and error with different training parameters. The most common technique to reduce training time is transfer learning [36]. With respect to traditional CV, the discrete Fourier transform is another CV technique which once experienced major popularity but now seems obscure. The algorithm can be used to speed up convolutions as demonstrated by [37, 38] and hence may again become of major importance.

However, it must be said that easier, more domain-specific tasks than general image classification will not require as much data (in the order of hundreds or thousands rather than millions). This is still a considerable amount of data and CV techniques are often used to boost training data through data augmentation or reduce the data down to a particular type of feature through other pre-processing steps.

Pre-processing entails transforming the data (usually with traditional CV techniques) to allow relationships/patterns to be more easily interpreted before training your model. Data augmentation is a common pre-processing task which is used when there is limited training data. It can involve performing random rotations, shifts, shears, etc. on the images in your training set to effectively increase the number of training images [19]. Another approach is to highlight features of interest before passing the data to a CNN with CV-based methods such as background subtraction and segmentation [39].

### 3.3 Making Best Use of Edge Computing

If algorithms and neural network inferences can be run at the edge, latency, costs, cloud storage and processing requirements, and bandwidth requirements are reduced compared to cloud-based implementations. Edge computing can also privacy and security requirements by avoiding transmission of sensitive or identifiable data over the network.

Hybrid or composite approaches involving conventional CV and DL take great advantage of the heterogeneous computing capabilities available at the edge. A heterogeneous compute architecture consists of a combination of CPUs, microcontroller coprocessors, Digital Signal Processors (DSPs), Field Programmable Gate Arrays (FPGAs) and AI accelerating devices [31] and can be power efficient by assigning different workloads to the most efficient compute engine. Test implementations show 10x latency reductions in object detection when DL inferences are executed on a DSP versus a CPU [28].

Several hybrids of deep learning and hand-crafted features based approaches have demonstrated their benefits in edge applications. For example, for facial-expression recognition, [41] propose a new feature loss to embed the information of hand-crafted features into the training process of network, which tries to reduce the difference between hand-crafted features and features learned by the deep neural network. The use of hybrid approaches has also been shown to be advantageous in incorporating data from other sensors on edge nodes. Such a hybrid model where the deep learning is assisted by additional sensor sources like synthetic aperture radar (SAR) imagery and elevation like synthetic aperture radar (SAR) imagery and elevation is presented by [40]. In the context of 3D robot vision, [42] have shown that combining both linear subspace methods and deep convolutional prediction achieves improved performance along with several orders of magnitude faster runtime performance compared to the state of the art.

### 3.4 Problems Not Suited to Deep Learning

There are many more changing problems in CV such as: Robotic, augmented reality, automatic panorama stitching, virtual reality, 3D modelling, motion stamation, video stabilization, motion capture, video processing and scene understanding which cannot simply be easily implemented in a differentiable manner with deep learning but need to be solved using the other "traditional" techniques.

DL excels at solving closed-end classification problems, in which a wide range of potential signals must be mapped onto a limited number of categories, given that there is enough data available and the test set closely resembles the training set. However, deviations from these assumptions can cause problems and it is critical to acknowledge the problems which DL is not good at. Marcus et al. present ten concerns for deep learning, and suggest that deep learning must be supplemented by other techniques if we are to reach artificial general intelligence [43]. As well as discussing the limitations of the training procedure and intense computing and data requirements as we do in our paper, key to their discussion is identifying problems where DL performs poorly and where it can be supplemented by other techniques.

One such problem is the limited ability of DL algorithms to learn visual relations, i.e. identifying whether multiple objects in an image are the same or different. This limitation has been demonstrated by [43] who argue that feedback mechanisms including attention and perceptual grouping may be the key computational components to realising abstract visual reasoning.

It is also worth noting that ML models find it difficult to deal with priors, that is, not everything can be learnt from data, so some priors must be injected into the models [44], [45]. Solutions that have to do with 3D CV need strong priors in order to work well, e.g. image-based 3D modelling requires smoothness, silhouette and illumination information [46].

Below are some emerging fields in CV where DL faces new challenges and where classic CV will have a more prominent role.

### 3.5 3D Vision

3D vision systems are becoming increasingly accessible and as such there has been a lot of progress in the design of 3D Convolutional Neural Networks (3D CNNs). This emerging field is known as Geometric Deep Learning and has multiple applications such as video classification, computer graphics, vision and robotics. This paper will focus on 3DCNNs for processing data from 3D Vision Systems. Wherein 2D convolutional layers the kernel has the same depth so as to output a 2D matrix, the depth of a 3D convolutional kernel must be less than that of the 3D input volume so that the output of the convolution is also 3D and so preserve the spatial information.

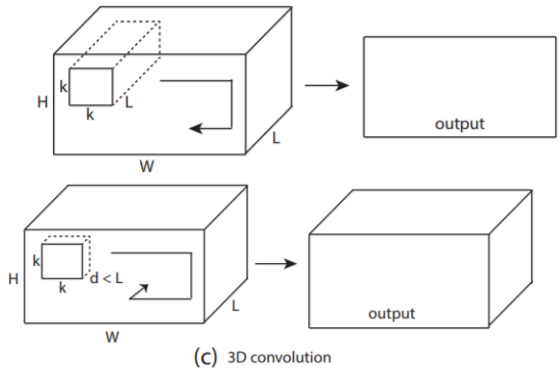

**Fig. 3.** 2DCNN vs. 3D CNN [47]

The size of the input is much larger in terms of memory than conventional RGB images and the kernel must also be convolved through the input space in 3 dimensions (see Fig. 3). As a result, the computational complexity of 3D CNNs grows cubically with resolution. Compared to 2D image processing, 3D CV is made even more difficult as the extra dimension introduces more uncertainties, such as occlusions and different cameras angles as shown in Fig. 4.

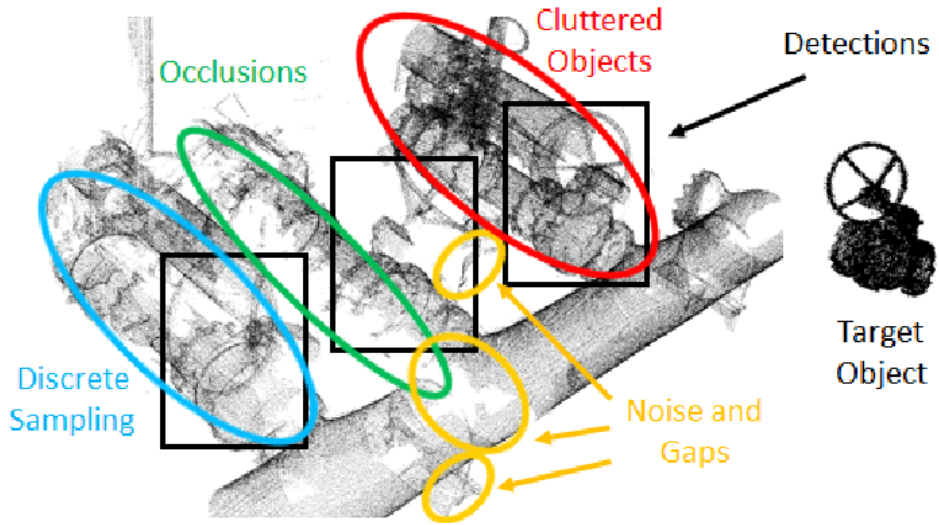

**Fig. 4.** 3D object detection in point clouds is a challenging problem due to discrete sampling, noisy scans, occlusions and cluttered scenes. Figure from [48].

FFT based methods can optimise 3D CNNs reduce the amount of computation, at the cost of increased memory requirements however. Recent research has seen the implementation of the Winograd Minimal Filtering Algorithm (WMFA) achieve a two-fold speedup compared to cuDNN (NVIDIA's language/API for programming on their graphics cards) without increasing the required memory [49]. The next section will include some solutions with novel architectures and pre-processing steps to various 3D data representations which have been proposed to overcome these challenges.

Geometric Deep Learning (GDL) deals with the extension of DL techniques to 3D data. 3D data can be represented in a variety of different ways which can be classified as Euclidean or non-Euclidean [50].3D Euclidean-structured data has an underlying grid structure that allows for a global parametrization and having a common system of coordinates as in 2D images. This allows existing 2D DL paradigms and 2DCNNs can be applied to 3D data. 3D Euclidean data is more suitable for analysing simple rigid objects such as, chairs, planes, etc e.g. with voxel-based approaches [51]. On the other hand, 3D non-Euclidean data do not have the gridded array structure where there is no global parametrization. Therefore, extending classical DL techniques to such representations is a challenging task and has only recently been realized with architectures such as Pointnet [52].

Continuous shape information that is useful for recognition is often lost in their conversion to a voxel representation. With respect to traditional CV algorithms, [53] propose a single dimensional feature that can be applied to voxel CNNs. A novel rotation-invariant feature based on mean curvature that improves shape recognition for voxel CNNs was proposed. The method was very successful in that when it was applied to the state-of-the-art recent voxel CNN Octnet architecture a 1% overall accuracy increase on the ModelNet10 dataset was achieved.

## 3.6 SLAM

Visual SLAM is a subset of SLAM where a vision system is used instead of LiDAR for the registration of landmarks in a scene. Visual SLAM has the advantages of photogrammetry (rich visual data, low-cost, lightweight and low power consumption) without the associated heavy computational workload involved in post-processing. The visual SLAM problem consists of steps such as environment sensing, data matching, motion estimation, as well as location update and registration of new landmarks [54].

Building a model of how visual objects appear in different conditions such as 3D rotation, scaling, lighting and extending from that representation using a strong form of transfer learning to achieve zero/one shot learning is a challenging problem in this domain. Feature extraction and data representation methods can be useful to reduce the amount of training examples needed for an ML model [55].

A two-step approach is commonly used in image based localization; place recognition followed by pose estimation. The former computes a global descriptor for each of the images by aggregating local image descriptors, e.g. SIFT, using the bag-of-words approach. Each global descriptor is stored in the database together with the camera pose of its associated image with respect to the 3D point cloud reference map. Similar global descriptors are extracted from the query image and the closest global descriptor in the database can be retrieved via an efficient search. The camera pose of the closest global descriptor would give us a coarse localization of the query image with respect to the reference map. In pose estimation, the exact pose of the query image calculated more precisely with algorithms such as the Perspective-n-Point (PnP) [13] and geometric verification [18] algorithms. [56]

The success of image based place recognition is largely attributed to the ability to extract image feature descriptors. Unfortunately, there is no algorithm to extract local features similar to SIFT for LiDAR scans. A 3D scene is composed of 3D points and database images. One approach has associated each 3D point to a set of SIFT descriptors corresponding to the image features from which the point was triangulated. These descriptors can then be averaged into a single SIFT descriptor that describes the appearance of that point [57].

Another approach constructs multi-modal features from RGB-D data rather than the depth processing. For the depth processing part, they adopt the well-known colourization method based on surface normals, since it has been proved to be effective and robust across tasks [58]. Another alternative approach utilizing traditional CV techniques presents the Force Histogram Decomposition (FHD), a graph-based hierarchical descriptor that allows the spatial relations and shape information between the pairwise structural subparts of objects to be characterized. An advantage of this learning procedure is its compatibility with traditional bags-of-features frameworks, allowing for hybrid representations gathering structural and local features [59].

## 3.7 360 cameras

A 360 camera, also known as an omnidirectional or spherical or panoramic camera is a camera with a 360-degree field of view in the horizontal plane, or with a visual field that covers (approximately) the entire sphere. Omnidirectional cameras are important in applications such as robotics where large visual field coverage is needed. A 360 camera can replace multiple monocular cameras and eliminate blind spots which obviously advantageous in omnidirectional Unmanned Ground Vehicles (UGVs) and Unmanned Aerial Vehicles (UAVs). Thanks to the imaging characteristic of spherical cameras, each image captures the 360◦ panorama of the scene, eliminating the limitation on available steering choices. One of the major challenges with spherical images is the heavy barrel distortion due to the ultra-wide-angle fisheye lens, which complicates the implementation of conventional human vision inspired methods such as lane detection and trajectory tracking. Additional pre-processing steps such as prior calibration and deworming are often required. An alternative approach which has been presented by [60], who circumvent these pre-processing steps by formulating navigation as a classification problem on finding the optimal potential path orientation directly based on the raw, uncalibrated spherical images.

Panorama stitching is another open research problem in this area. A real-time stitching methodology [61] uses a group of deformable meshes and the final image and combine the inputs using a robust pixel-shader. Another approach [62], combine the accuracy provided by geometric reasoning (lines and vanishing points) with the higher level of data abstraction and pattern recognition achieved by DL techniques (edge and normal maps) to extract structural and generate layout hypotheses for indoor scenes. In sparsely structured scenes, feature-based image alignment methods often fail due to shortage of distinct image features. Instead, direct image alignment methods, such as those based on phase correlation, can be applied. Correlation-based image alignment techniques based on discriminative correlation filters (DCF) have been investigated by [23] who show that the proposed DCF-based methods outperform phase correlation-based approaches on these datasets.

### 3.8 Dataset Annotation and Augmentation

There are arguments against the combination of CV and DL and they summarize to the conclusion that we need to re-evaluate our methods from rule-based to data-driven. Traditionally, from the perspective of signal processing, we know the operational connotations of CV algorithms such as SIFT and SURF methods, but DL leads such meaning nowhere, all you need is more data. This can be seen as a huge step forward, but may be also a backward move. Some of the pros and cons of each side of this debate have been discussed already in this paper, however, if future-methods are to be purely data-driven then focus should be placed on more intelligent methods for dataset creation.

The fundamental problem of current research is that there is no longer enough data for advanced algorithms or models for special applications. Coupling custom datasets and DL models will be the future theme to many research papers. So many researchers' outputs consist of not only algorithms or architectures, but also datasets or methods to amass data. Dataset annotation is a major bottleneck in the DL workflow which requires many hours of manual labelling. Nowhere is this more problematic than in semantic segmentation applications where every pixel needs to be annotated accurately. There are many useful tools available to semi-automate the process as reviewed by [20], many of which take advantage of algorithmic approaches such as ORB features [55], polygon morphing [63], semi-automatic Area of Interest (AOI) fitting [55] and all of the above [63].

The easiest and most common method to overcome limited datasets and reduce overfitting of deep learning models for image classification is to artificially enlarge the dataset using label-preserving transformations. This process is known as dataset augmentation and it involves the artificial generation of extra training data from the available ones, for example, by cropping, scaling, or rotating images [64]. It is desirable for data augmentation procedures to require very little computation and to be implementable within the DL training pipeline so that the transformed images do not need to be stored on disk. Traditional algorithmic approaches that have been employed for dataset augmentation include Principle Component Analysis (PCA) [1], adding noise, interpolating or extrapolating between samples in a feature space [65] and modelling the visual context surrounding objects from segmentation annotations [66].

# Conclusion

A lot of the CV techniques invented in the past 20 years have become irrelevant in recent years because of DL. However, knowledge is never obsolete and there is always something worth learning from each generation of innovation. That knowledge can give you more intuitions and tools to use especially when you wish to deal with 3D CV problems for example. Knowing only DL for CV will dramatically limit the kind of solutions in a CV engineer's arsenal.

In this paper we have laid down many arguments for why traditional CV techniques are still very much useful even in the age of DL. We have compared and contrasted traditional CV and DL for typical applications and discussed how sometimes traditional CV can be considered as an alternative in situations where DL is overkill for a specific task.

The paper also highlighted some areas where traditional CV techniques remain relevant such as being utilized in hybrid approaches to improve performance. DL innovations are driving exciting breakthroughs for the IoT (Internet of Things), as well as hybrid techniques that combine the technologies with traditional algorithms. Additionally, we reviewed how traditional CV techniques can actually improve DL performance in a wide range of applications from reducing training time, processing and data requirements to being applied in emerging fields such as SLAM, Panoramic-stitching, Geometric Deep Learning and 3D vision where DL is not yet well established.

The digital image processing domain has undergone some very dramatic changes recently and in a very short period. So much so it has led us to question whether the CV techniques that were in vogue prior to the AI explosion are still relevant. This paper hopefully highlight some cases where traditional CV techniques are useful and that there is something still to gain from the years of effort put in to their development even in the age of data-driven intelligence.